\documentclass[conference]{IEEEtran}
\IEEEoverridecommandlockouts

\usepackage{cite}
\usepackage{amsmath,amssymb,amsfonts}
\usepackage{graphicx}
\usepackage{textcomp}
\usepackage{xcolor}
\usepackage{booktabs}
\usepackage{siunitx}
\usepackage{bbm}
\usepackage{mathrsfs}
\usepackage{float}
\usepackage{colortbl}
\usepackage{algpseudocode}
\usepackage{bm}

\usepackage{enumitem}
\usepackage{threeparttable}
\usepackage{placeins}
\usepackage{pifont}

\graphicspath{{imgs/}}

\setlist[enumerate,1]{itemsep=3pt}

\definecolor{formalgreen}{rgb}{0.1, 0.7, 0.1}
\definecolor{formalred}{rgb}{0.9, 0.2, 0.2}

\begin{document}

%\title{Traversing Narrow Paths: A Two-Stage Reinforcement Learning Framework for Robust and Safe Humanoid Walking}
\title{Physics-Guided Residual Reinforcement Learning for Humanoid Narrow-Path Traversal%
\thanks{This work was supported in part by the National Natural Science Foundation of China under Grant U22B2040, in part by the Major Project of Anhui Province's Science and Technology Innovation Breakthrough Plan (202423h08050003), and in part by the Key Science \& Technology Project of Anhui Province (202523j08050001).}
}

% Long author format is supported by the bundled IEEEtran_HOWTO.pdf, Sec. IV-B2.
% Preserve author order; do not regroup authors by institution.
\author{
\IEEEauthorblockN{
Tianchen Huang\textsuperscript{1,\textdagger},
Sisheng Chen\textsuperscript{1,\textdagger},
Wei Zhou\textsuperscript{1},
Haopeng Zhang\textsuperscript{1},
Jiarong Sun\textsuperscript{1},
Ya Wang\textsuperscript{2},
Yumin Wang\textsuperscript{2},\\
Deguang Lyu\textsuperscript{2},
Yu Wang\textsuperscript{1,*},
Wei Gao\textsuperscript{1,*},
and Shiwu Zhang\textsuperscript{1}
}

\IEEEauthorblockA{
\textsuperscript{1}\textit{Institute of Humanoid Robots,
Department of Precision Machinery and Precision Instrumentation}\\
\textit{University of Science and Technology of China},
Hefei, Anhui 230026, China
}

\IEEEauthorblockA{
\textsuperscript{2}\textit{Openmind (Wuhu) Intelligent Robotics Co., Ltd.}\\
Wuhu, Anhui 241000, China
}

\IEEEauthorblockA{
\textsuperscript{\textdagger}These authors contributed equally
to this work and share first authorship.
}

\IEEEauthorblockA{
\textsuperscript{*}Corresponding authors:
wangyuustc@ustc.edu.cn; weigao@ustc.edu.cn
}
}

\maketitle

\begin{abstract}
% Traversing narrow paths with sparse, safety–critical contacts is challenging for humanoids and exposes the fragility of purely learned policies. We propose a physically grounded, two–stage framework that couples an LIP footstep template with a lightweight residual planner and a simple low–level tracker. \emph{Stage–I} is trained on flat ground: the tracker learns to robustly follow footstep targets by adding small random perturbations to the LIP template footsteps—without any hand–crafted centerline locking—thereby acquiring stable contact scheduling and strong target–tracking robustness. \emph{Stage–II} is trained in narrow–path simulation (instantiated in our experiments by a straight beam): a high–level planner predicts a body–frame residual $(\Delta x,\Delta y,\Delta \psi)$ for the swing foot only, refining the template step to prioritize safe, precise placement under narrow support while preserving interpretability. To ease deployment, sensing is kept minimal and sim–real consistent: the planner consumes compact, forward–facing elevation cues together with onboard IMU/joint signals. As an evaluation instantiation, we validate on a Unitree G1 traversing a $0.20$\,m–wide, $3$\,m–long beam. Across simulation and real–world studies, residual refinement consistently outperforms template–only and monolithic baselines in success rate, centerline adherence, and safety margins, while the structured footstep interface enables transparent analysis and low–friction sim–to–real transfer.
Traversing narrow paths is challenging for humanoid robots due to the sparse and safety-critical footholds required. Purely template-based or end-to-end reinforcement learning-based methods suffer from such harsh terrains. This paper proposes a two–stage training framework for such narrow path traversing tasks, coupling a template-based foothold planner with a low-level foothold tracker from Stage-I training and a lightweight perception-aided foothold modifier from Stage-II training. With the curriculum setup from flat ground to narrow paths across stages, the resulting controller in turn learns to robustly track and safely modify foothold targets to ensure precise foot placement over narrow paths. This framework preserves the interpretability from the physics-based template and takes advantage of the generalization capability from reinforcement learning, resulting in easy sim-to-real transfer. 
% The learned policies are validated on a Unitree G1 humanoid robot, leading to successful traversal of a 0.2~m-wide and 3~m-long beam for 20 trials without any failure. 
% This outperforms purely template-based or reinforcement learning-based baselines in terms of success rate, centerline adherence and safety margins.
The learned policies outperform purely template-based or reinforcement learning-based baselines in terms of success rate, centerline adherence and safety margins. Validation on a Unitree G1 humanoid robot yields successful traversal of a 0.2~m-wide and 3~m-long beam for 20 trials without any failure.
\end{abstract}


% Keep first-page figures from floating above the abstract.
\suppressfloats[t]

\section{Introduction}
% Humanoid traversal of narrow paths is safety–critical: feasible footholds shrink to centimeters, recovery margins vanish, and even modest perception or control delays can precipitate failure. Success therefore hinges on precise and interpretable footstep decisions, robust swing–stance execution, and low–friction sim–to–real deployment under minimal sensing.

Safe and accurate footholds are critical for humanoid robots traversing narrow paths, where the path width for feasible footholds shrink to one foot level and even modest perception or control delays can precipitate failure due to vanished recovery margins. Therefore, successful narrow path traversal hinges on efficient terrain perception, precise foothold selection and robust locomotion control.

\begin{figure}[!t] 
    \centering
    \includegraphics[width=0.8\columnwidth]{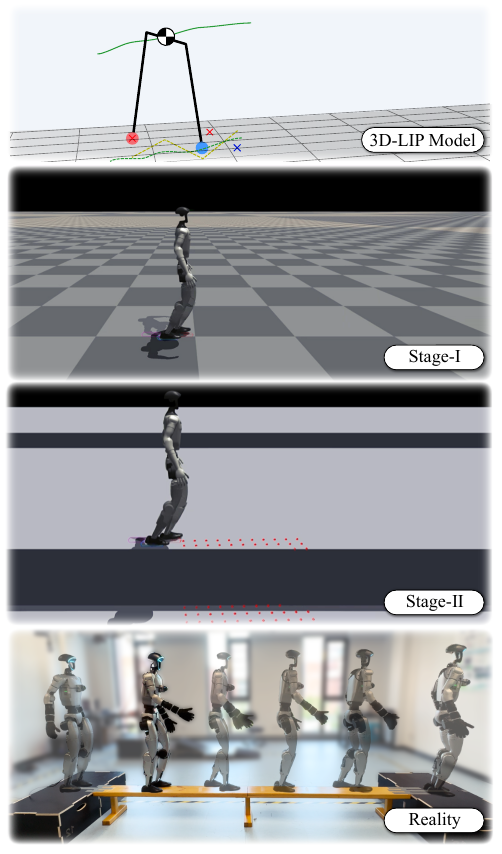}
    % \vspace{-1pt}
    \caption{Overview. The proposed framework uses 3D-LIPM as foothold planner, low-level policy from Stage-I training as foothold tracker and high-level policy from Stage-II training as foothold modifier for robust and safe narrow path walking. Successful experimental validation on the Unitree G1 humanoid robot is performed.}
    \label{fig:sim2real}
    % \vspace{-15pt}
\end{figure}

% Existing approaches largely follow two complementary threads. On one side, physics–grounded stepping rules—most notably the \emph{Linear Inverted Pendulum} (LIP) and its \emph{Extrapolated Center of Mass} (XCoM) capture–point rule—provide compact, predictive templates for where to place the next foot, shaping both biomechanics and robotic balance control. On the other, residual/hybrid learning augments a nominal controller with data–driven refinements, improving robustness without discarding a model–based scaffold~\cite{Lee2024Integrating}. For sparse–foothold settings, recent results emphasize curricula and lightweight exteroception, and coupling template footstep planning with model–free RL has yielded hardware–validated gains~\cite{BeamDojo2025}. 

Within this context, existing approaches mainly follow two paradigms. On the one hand, model-based foothold generation methods
% , most notably the \emph{Instantaneous Capture Point} (ICP) method based on the \emph{Linear Inverted Pendulum} (LIP) model, 
take advantage of compact template models to provide guidance for where to place the swing foot to yield balanced locomotion control~\cite{Pratt2006CapturePoint}. On the other hand, model-free Reinforcement Learning (RL) methods learn end-to-end foothold selection and locomotion control from data, \emph{e.g.}, an attention-based terrain map encoder trained jointly with the control policy has realized generalized legged locomotion~\cite{he2025attention}. When given sparse foothold options, learning-based methods typically introduce curriculum schedules and lightweight exteroception to cope with sparse rewards. A representative example is \emph{BeamDojo}, which employs a two-stage RL pipeline with a sampling-based foothold reward and onboard LiDAR terrain height mapping to achieve hardware-validated traversal over narrow beams and stepping stones~\cite{BeamDojo2025}.
% For scenarios requiring sparse footholds, recent studies emphasize curricula setup and lightweight exteroception under the coupling of template-based footstep planning with model–free Reinforcement Learning (RL). This combination has yielded hardware–validated performance improvements~\cite{BeamDojo2025}. 

% Despite this progress, narrow–path traversal remains demanding. Purely end–to–end RL methods can overfit simulator–specific assumptions, faces sparse/unsafe exploration, and offers limited interpretability in safety–critical regimes. Conversely, purely model–based templates are vulnerable to modeling errors, contact uncertainty, unmodeled compliance, and latency, degrading placement accuracy on sparse supports. These complementary limitations motivate a synthesis: retain a physics prior as an interpretable stepping scaffold and allocate learning to a small, safety–relevant interface. This perspective leads directly to the residual–over–template, two–stage framework developed next.

Despite the progress, narrow path traversal remains challenging. Purely end–to–end RL methods can overfit simulator–specific assumptions, facing unsafe exploration and suffering from limited foothold interpretability in safety–critical tasks. Conversely, purely template–based methods are vulnerable to modeling discrepancy, contact uncertainty and control system latency, resulting in inaccurate foot placement on limited support areas. These limitations have motivated a category of synthesized methods, which retain a physics-based prior as an interpretable foothold planner and allocate residual learning as a safety–relevant modifier. Thus, 
% recent studies emphasize curricula setup and lightweight exteroception under the coupling of template-based footstep planning with model–free RL, where 
the residual learning methods can augment the nominal controller with data–driven refinements, improving robustness without discarding insights from physical models~\cite{Lee2024Integrating}. 
% For scenarios requiring sparse footholds, this combination has yielded hardware–validated performance improvements~\cite{BeamDojo2025}.
% This perspective leads directly to the residual–over–template, two–stage framework developed next.

% \textbf{Our approach.} We propose a physically grounded, two–stage reinforcement learning framework for narrow path traversing that couples a LIP footstep template with a lightweight residual planner and a simple low–level tracker. \emph{Stage–1} is trained on flat ground: we intentionally add small random perturbations to the LIP template footstep planning during training so that the tracker learns robust foothold following and stable contact scheduling—\emph{without} any hand–crafted centerline locking. \emph{Stage–2} is trained in simulation on narrow paths substantiated by a narrow beam: a high–level planner predicts a body–frame residual $(\Delta x,\Delta y,\Delta \psi)$ for the \emph{swing} foot only, refining the template step to emphasize safe, precise placements under narrow support while preserving interpretability. To ease deployment, sensing is kept minimal and consistent across simulation and physical robots: the planner consumes compact, forward–facing elevation cues, as well as onboard IMU/joint signals, avoiding heavyweight vision pipelines while retaining the information needed for safe foot placement.

Following this flavor, this paper proposes a lightweight and efficient two–stage reinforcement learning framework for robust and safe narrow path traversal.
Stage-I is trained in simulation on flat ground: a Linear Inverted Pendulum model (LIPM) based foothold planner is utilized during training such that the low-level RL-based tracker can robustly follow each foothold and realize stable contact scheduling. 
Stage-II is trained in simulation on narrow paths: a high–level RL-based modifier generates a body–frame residual for the swing foot only, refining the foothold generated by the planner to ensure safe and precise foot placements on narrow paths. This setup helps preserve the interpretability from physics-based template models. Additionally, in the proposed method, sensing is kept minimal and consistent across the simulated and the physical robots to ease deployment, 
% the modifier consumes compact, forward–facing elevation cues, as well as onboard IMU/joint signals, 
avoiding heavyweight vision pipelines with only necessary information for safe foot placement.

% Overall, we make three key contributions:
% \begin{enumerate}
% \item \emph{Physics-guided residual stepping with a two-stage curriculum.} We refine a LIP footstep template using a bounded, body-frame residual $(\Delta x,\Delta y,\Delta\psi)$ applied to the swing foot at step transitions (event-driven, zero-order hold); Stage–1 learns a robust tracker via disturbance–target training on flat ground, and Stage–2 optimizes path-aware objectives on narrow supports.
% \item \emph{Minimal sensing \& hardware proof.} Using only compact forward elevation cues plus onboard IMU/joint signals with a sim–real consistent representation, a Unitree~G1 reliably traverses a $0.2$\,m$\times$3\,m beam, outperforming template–only and monolithic baselines.
% \end{enumerate}

Overall, this paper makes two key contributions:
\begin{enumerate}
\item \emph{Physics-guided foot placement learning with a two-stage training curriculum for narrow path traversal.} The LIPM-based foothold is refined by the bounded body-frame residual to achieve robust and safe foothold selection for the swing leg. Stage-I learns a robust foothold tracker via intentionally added target disturbance through training on flat ground, and Stage-II optimizes terrain-aware objectives for a safe foothold modifier on narrow paths.
\item \emph{Minimal sensing requirement and experimental proof on a physical humanoid robot.} Using only compact anterior terrain height sampling maps and onboard IMU/joint signals with consistent representation in both simulation and experiments, a Unitree G1 humanoid robot has been able to reliably traverse a narrow beam of $0.2\,m$ wide and $3\,m$ long, outperforming methods based on either template models or Reinforcement Learning purely.
\end{enumerate}

\section{Related Work}
\subsection{Physics-Based Foothold Planning for Locomotion Control}

% Early biped locomotion formalized balance and stepping via simplified models and analytic stability measures. Preview control of the Zero-Moment Point (ZMP) with the Linear Inverted Pendulum model enabled pattern generation that respects balance constraints \cite{Kajita2003Preview}. The \emph{extrapolated center of mass} concept connected CoM state to required foot placement for recovery \cite{Hof2008XCoM}, and related \emph{capture point} ideas framed “when and where to step” for push recovery \cite{Pratt2006CapturePoint}. Building on these foundations, N-step \emph{capturability} provided a theoretical lens on feasible stabilization with a finite number of steps \cite{Koolen2012Capturability}. These physics-grounded templates remain influential because they yield interpretable balance rules and compact footstep representations.

Bipedal robots from early years achieve balanced walking via reduced-order models and analytic stability measures. Preview control of Zero-Moment Point (ZMP) based on the Linear Inverted Pendulum model enables locomotion pattern generation that respects balance constraints~\cite{Kajita2003Preview}. N-step capturability has provided a theoretical framework for feasible stabilization within a finite number of steps~\cite{Koolen2012Capturability}. The extrapolated Center of Mass (CoM) concept connects CoM state to required foot placement for balance control~\cite{Hof2008XCoM}, and has led to the idea of Instantaneous Capture Point (ICP) for selecting “when and where to step” for push recovery~\cite{Pratt2006CapturePoint}. These physics-grounded methods remain influential because they yield interpretable balance rules and compact foothold representations.

% Within template-based locomotion, Model Predictive Control (MPC) has been used to \emph{explicitly optimize future footstep locations} together with CoM dynamics, enabling online adaptation to pushes and modeling errors. Seminal LIPM–MPC formulations include trajectory-free linear MPC with \emph{online footstep adjustment}~\cite{Wieber2006MPC} (online step placement under strong perturbations), adaptive foot positioning through linear MPC~\cite{Diedam2008LMPC}, and automatic footstep placement via MPC that tracks a desired velocity while adjusting footsteps~\cite{Herdt2010MPC}. Recent variants plan both step position and orientation or leverage reduced-order models such as DCM/ALIP for improved prediction and feasibility handling~\cite{Ding2022OrientationMPC,Griffin2019DCMMPC,Gibson2021ALIPMPC,Acosta2023ConstrainedFootholds}.

With model-based foothold planners, Model Predictive Control (MPC) has been extensively used to explicitly optimize future footholds under CoM dynamics, enabling online adaptation to external disturbances. The LIPM-MPC formulations take advantage of linear MPC to adjust future footholds online, for either balance control or velocity tracking~\cite{Wieber2006MPC,Diedam2008LMPC,Herdt2010MPC}.
% The LIPM–MPC formulations include trajectory-free linear MPC with \emph{online footstep adjustment}~\cite{Wieber2006MPC} (online step placement under strong perturbations), adaptive foot positioning through linear MPC~\cite{Diedam2008LMPC}, and automatic footstep placement via MPC that tracks a desired velocity while adjusting footsteps~\cite{Herdt2010MPC}. 
Recent variants can plan both step position and orientation~\cite{Ding2022OrientationMPC,Acosta2023ConstrainedFootholds}, or leverage reduced-order models such as DCM/ALIP, for improved locomotion performance~\cite{Griffin2019DCMMPC,Gibson2021ALIPMPC}.
Additionally, adapting step duration can also markedly improve landing accuracy for restricted footholds. By modulating swing duration online, precise contacts can be realized even when feasible regions are small~\cite{AdaptiveStepDuration}. 

% Orthogonal to spatial footstep selection, adapting the \emph{step duration} can markedly improve landing accuracy on \emph{restricted footholds}. By modulating swing timing online, precise contacts can be realized even when feasible regions are small, which is directly relevant to beam/stepping-stone tasks \cite{AdaptiveStepDuration}. Such timing adaptation is complementary to ALIP-based templates and can be integrated with MPC-based pipelines.

% Bridging model structure with learning, recent work combines template footstep planning with model-free RL, using physics guidance to shape the action space while retaining adaptability \cite{Lee2024Integrating}. That line shows hardware-validated improvements when an RL policy tracks template-suggested footsteps rather than full trajectories. However, that approach is tailored to flat-ground locomotion and does not incorporate exteroceptive terrain perception or perception-driven footstep adjustment, thus it does not address sparse-support tasks such as beam traversal. In contrast, we retain an interpretable LIP template, and confine learning to a lightweight swing-foot residual, with a two-stage curriculum that first builds a robust target-tracking controller on flat ground and then learns beam-specific refinements under minimal, sim-real consistent sensing.

On the other hand, recent work also combines template-based foothold planner with model-free RL, using physics-based guidance to shape the action space during training~\cite{Lee2024Integrating}. Hardware-validated improvements have been reported with RL policies tracking template-based footholds. However, this approach so far is only tailored to flat-ground locomotion.
When sparse-support tasks such as beam traversal are confronted, recent advancements have proposed to incorporate exteroceptive terrain perception for foothold adjustment within purely RL framework~\cite{BeamDojo2025,he2025attention}. Nevertheless, an efficient narrow path traversing control framework with high success rate and careful safety consideration based on interpretable physical models is still lacking. Therefore, this paper utilizes the Linear Inverted Pendulum model, and confines learning to a lightweight residual learning. 
% With a two-stage curriculum, the proposed method first trains a robust foothold tracker on flat ground and then a safe foothold modifier on narrow beams under minimal terrain perception.

\subsection{Residual Learning with Staged Curriculum and Local Terrain Perception}

% Residual learning approaches augment a nominal controller or template with a learned correction, improving performance while preserving structure and interpretability. In manipulation and control, \emph{Residual Policy Learning} improves nondifferentiable controllers by learning a residual on top of them \cite{Silver2018RPL}; similarly, \emph{Residual Reinforcement Learning} demonstrates that decomposing a task into a hand-designed component plus a learned residual yields data-efficient, hardware-validated behaviors \cite{Johannink2019ResidualRL}. However, a residual without exteroceptive terrain cues cannot reason about restricted footholds; on narrow paths the supervision collapses to sparse, high-risk rewards. This motivates coupling the residual interface with lightweight local elevation sensing and staged training for sparse supports—topics we review next.

Residual learning approaches augment a nominal controller with a learned correction, improving performance while preserving attributes of the nominal controller. In manipulation, Residual Policy Learning improves nondifferentiable policies by learning an additional residual policy~\cite{Silver2018RPL}. Similarly, Residual Reinforcement Learning demonstrates that decomposing a controller into a physics-based model and a learned residual policy yields data-efficient behaviors on hardware~\cite{Johannink2019ResidualRL}. 

% Sparse footholds exacerbate reward sparsity and failure sensitivity, making end-to-end learning brittle without additional structure or curricula. In computer animation and robotics, \emph{ALLSTEPS} showed that curriculum-driven RL can master stepping-stone locomotion and highlighted the importance of staged learning for contact-constrained tasks \cite{Xie2020ALLSTEPS}. On hardware, Cassie demonstrated closed-loop walking over stepping stones using learning-based methods, underscoring the need for precise foot placement under constrained supports \cite{Duan2022SteppingStones}. Most pertinent to our setting, \emph{BeamDojo} tackles humanoid locomotion on sparse footholds with a two-stage RL pipeline and LiDAR-based elevation mapping, reporting both simulation and real-world success on beams and stepping stones \cite{BeamDojo2025}. Our formulation shares the staged philosophy but differs by retaining an explicit LIP template and restricting learning to a lightweight residual interface tailored to sparse-contact safety.

However, for the narrow path traversing tasks considered in this paper, the supervision for residual learning can easily collapse due to sparse and high-risk foothold rewards. The failure sensitivity of the training process makes residual learning brittle without additional framework structure or training curricula. In simulation, \emph{ALLSTEPS} has showed that curriculum-driven RL can master stepping-stone locomotion and highlighted the importance of staged learning for contact-constrained locomotion tasks~\cite{Xie2020ALLSTEPS}. 
% On hardware, the bipedal robot Cassie demonstrated closed-loop walking control over stepping stones using learning-based methods, underscoring the need for precise foot placement under constrained terrain supports~\cite{Duan2022SteppingStones}. 
Most pertinent to our settings, \emph{BeamDojo} tackles humanoid locomotion with sparse footholds using a two-stage RL pipeline based on LiDAR-based terrain height sampling maps, resulting in successful locomotion on beams and stepping stones in both simulated and real worlds~\cite{BeamDojo2025}. Therefore, the proposed formulation in this paper shares the staged training philosophy but differs by using the explicit LIPM as foothold planner and the lightweight residual learning as foothold modifier to realize more robust and safer locomotion behaviors.

% Local terrain height maps have emerged as an effective exteroceptive representation for adequate foot placement. Robust quadruped controllers trained in simulation succeed in natural environments by consuming compact height sample maps rather than heavy vision stacks. Efficient GPU elevation mapping pipelines further enable real-time, robot-centric height grids for locomotion \cite{Miki2022ElevationGPU}. Recent perceptive locomotion systems integrate such local elevation maps with proprioception to achieve field robustness \cite{Miki2022WildANYmal}, while massively parallel RL and task curricula accelerate policy training for contact-rich mobility \cite{Rudin2022Minutes}. High-agility demonstrations (e.g., ANYmal Parkour) further validate the viability of compact terrain representations in hardware \cite{Hoeller2024Parkour}. We follow this trend by adopting a minimal, forward-facing elevation window that remains consistent between simulation and the real robot, enabling our residual-over-template planner to refine footsteps for narrow-beam traversal without reliance on heavyweight perception.

As mentioned, to determine adequate footholds through residual learning, local terrain height sampling maps have emerged as a necessary part of the framework. Robust quadruped controllers trained in simulation have succeeded in natural environments by consuming compact height maps rather than heavy vision stacks. Efficient terrain height mapping pipelines via Graphics Processing Units further enable real-time robot-centric terrain height maps for locomotion control~\cite{Miki2022ElevationGPU}. Recent progress integrates such local terrain height maps to achieve field robustness~\cite{Miki2022WildANYmal}. High-agility behavior demonstrations (\emph{e.g.}, ANYmal Parkour) further validate the viability of compact terrain perception on hardware \cite{Hoeller2024Parkour}. This paper follows this trend by sampling a minimal anterior terrain height map, enabling the proposed controller to refine footholds for narrow path traversal without heavyweight perception.

\section{Method\label{sec:method}}
\begin{figure*}[!t] 
    \centering
    \includegraphics[width=\textwidth]{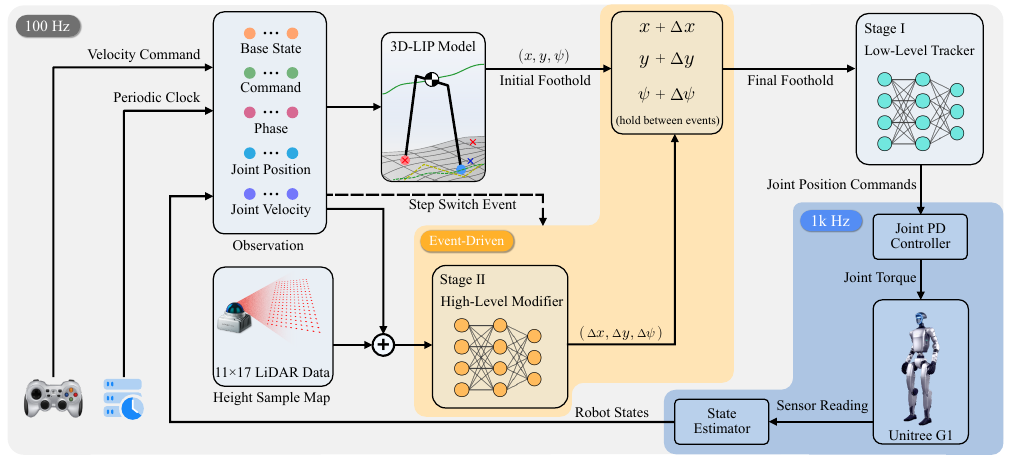}
    % \vspace{-1pt}
    % \caption{This figure illustrates our two-stage humanoid beam traversal framework. The system combines a LIP footstep template with a residual refinement for the swing foot. The low-level tracker operates at a high frequency of 1000 Hz (PD control loop) to track footstep targets, while the high-level residual planner updates at a slower rate of 100 Hz during step transitions in Stage-I. In Stage-II, the planner updates only when a step transition occurs. Stage-I employs disturbance-target training on flat ground to robustly learn footstep tracking, and the entire system operates asynchronously, with the planner and tracker phase-synchronized to ensure safe and precise foot placements on narrow beams.}
    \caption{The proposed framework for humanoid robot traversing narrow paths. A two-stage training curriculum is designed for the low-level foothold tracker and high-level foothold modifier. Different background colors indicate different operational frequencies}
    \label{fig:overview}
    % \vspace{-15pt}
\end{figure*}

\subsection{Framework Overview}

% Our goal is to enable safe and repeatable humanoid traversal on narrow beams, where contacts are sparse and failure is catastrophic. We address this with a lightweight, physically grounded architecture that separates \emph{where to step} from \emph{how to realize the step} (Fig.~\ref{fig:overview}). A model-based footstep template provides interpretable nominal contacts, a high-level policy contributes only a small correction, and a simple low-level tracker executes the resulting target with high frequency.

To enable safe and repeatable humanoid traversal over narrow paths like beams, where contacts are sparse and any failure can be catastrophic, this paper focuses on a lightweight and physics-based architecture that separates \emph{where to step} from \emph{how to realize the step}, as shown in Fig.~\ref{fig:overview}. 
% \textbf{Template $\rightarrow$ residual $\rightarrow$ tracking.} Given the robot state, a LIP-based template proposes the next \emph{swing-foot} target pose $u_{\text{temp}}$. The planner then predicts a body-frame residual $r \!=\! (\Delta x,\Delta y,\Delta \psi)$ that refines only this swing foot, producing the final target $u_{\text{final}} \!=\! u_{\text{temp}} \oplus r$, where $\oplus$ denotes composition in the footstep pose space. The low-level controller tracks $u_{\text{final}}$ by issuing joint position targets to the robot’s PD servos. This interface confines learning to a 3D, safety-relevant channel while keeping the nominal stepping physics explicit and interpretable.
Given the robot's states, a 3D-LIPM first plans the next swing-leg foothold target $u_{\text{init}}$. The high-level modifier then predicts a body-frame residual $\Delta u \!=\! (\Delta x,\Delta y,\Delta \psi)$ to refine the initial foothold as the final target $u_{\text{final}} \!=\! u_{\text{init}} \oplus \Delta u$, where $\oplus$ denotes composition in the task space. The low-level tracker finally issues desired joint positions to a Proportional-Derivative (PD) controller to ensure $u_{\text{final}}$. This framework confines learning to a safety-relevant role while keeps the nominal stepping physics explicit and interpretable.

% \textbf{Two-stage learning.} We train the two components with distinct objectives and an explicit timing scheme. (i) \emph{Stage–I} learns a robust tracker via disturbance–target training: at every step we add small, zero-mean random offsets to the LIP \emph{swing-foot} target so the policy learns reliable footstep tracking and clean swing–stance transitions under a fixed right/left alternation, without any hand-crafted centerline locking. During Stage–I rollouts the tracking policy runs at \textbf{100\,Hz} and outputs desired joint positions, which are executed by a joint PD loop at \textbf{1\,kHz}; no high-level planner is used. (ii) \emph{Stage–II} trains the residual planner to refine template steps on narrow support: the planner predicts a body-frame residual $(\Delta x,\Delta y,\Delta \psi)$ for the swing foot only. The planner is event-driven—queried once at each step transition—and its output is \emph{zero-order held} between events. During Stage–II rollouts and at deployment the timing remains the same: the tracker runs at \textbf{100\,Hz} with PD at \textbf{1\,kHz}, while the planner updates on step switches. The two policies thus run at different rates but are event-synchronized by the shared gait phase.

The foothold modifier and tracker are trained through RL with distinct objectives. \emph{Stage-I} trains the robust tracker via intentionally added foothold disturbances: a small and random zero-mean offset is added to the LIPM-based foothold target at every step, so the policy learns reliable foothold tracking and clean swing–stance transition under a scheduled right/left stance alternation, without any manual centerline locking. During Stage-I rollouts, the tracking policy runs at $100\,Hz$ and outputs desired joint positions, which are executed by the joint PD controller at $1\,kHz$. Note that the high-level modifier is not used in this stage. \emph{Stage-II} trains the foothold modifier to refine the initially planned footholds on narrow support: the modifier predicts a body-frame residual for the swing foot only. The modifier is event-driven, queried once when the step transition occurs. Its output is held constant between consecutive events. During Stage-II rollouts, the operational frequencies remain the same: the tracker runs at $100\,Hz$ with joint PD control at $1\,kHz$, while the modifier updates upon step transition events. The two policies thus run at different rates but are synchronized by the shared gait event.

% \textbf{Minimal sensing and sim-real consistency.} To ease deployment, the planner consumes only compact, forward-facing elevation cues together with onboard IMU and joint states; the same representation is used in both simulation and physical experiments. No heavy-weight vision stack is required. Policies are exported as TorchScript and deployed on a Unitree G1 humanoid robot with standard safety guards (torque limits, fall/edge detectors and foot–foot clearance checks). This design yields a thin, interpretable footstep interface, efficient training, and low-friction sim-to-real transfer for beam traversal.

To ease deployment, besides the onboard IMU signals and joint states, the control framework consumes only compact perception information. The same representation is used in both simulation and physical experiments. No heavy-weight vision stack is required. 
% Policies are exported as TorchScript and deployed on a Unitree G1 humanoid robot with standard safety guards, like torque limits, fall/edge detectors and foot–to-foot clearance checks. 
This results in a lightweight and interpretable framework for efficient training and sim-to-real transfer.

\subsection{Stage-I Training for Robust Foothold Tracker}

% We generate nominal footsteps with a lightweight, physics-based template and use Stage–I to turn them into a controller that robustly tracks nearby targets on flat ground. The template encodes balance structure while exposing a low-dimensional stepping interface; Stage–I then endows the low-level policy with invariance to small target variations, preparing it for narrow-support tasks.

Stage-I trains a robust low-level foothold tracker on flat ground that can adapt to small random foothold disturbances, preparing for the modified foothold on narrow paths by the residual from Stage-II policy. However, before the Stage-I training can be carried out, a model-based foothold planner has to be established first.  

% \paragraph*{Footstep template (LIP).} Assuming a constant CoM height $z_0$, the linear inverted pendulum yields $\omega_0=\sqrt{g/z_0}$ and the \emph{extrapolated CoM} $\xi_x = x + \dot{x}/\omega_0,\ \xi_y = y + \dot{y}/\omega_0$. At each step transition (phase threshold), the template proposes the next \emph{swing-foot} target $u_{\text{init}}=\Pi(\xi, v_{\text{cmd}}, \dot{\psi}_{\text{cmd}}, \text{phase})$, following capture-point for forward and heading updates. Only the swing foot is updated; the stance foot remains fixed until the next transition. This keeps nominal stepping physics explicit without task-specific hard constraints.Here $(x,y)$ and $(\dot{x},\dot{y})$ denote the current CoM position and velocity, $\xi=(\xi_x,\xi_y)$ is the Instantaneous Capture Point (ICP), and $\Pi(\cdot)$ maps the ICP, commanded velocity $(v_{\text{cmd}}, \dot{\psi}_{\text{cmd}})$, and gait phase to the next swing-foot target $u_{\text{init}}$.

\subsubsection{LIPM-based foothold planner}
Assuming a constant CoM height $z_0$, the LIP model yields $\omega_0=\sqrt{g/z_0}$, and then the Instantaneous Capture Point as $\xi_x = x + \dot{x}/\omega_0$ and $\xi_y = y + \dot{y}/\omega_0$. When step transition occurs, the planner proposes the next foothold target $u_{\text{init}}=\Pi(\xi, v_{\text{cmd}}, \dot{\psi}_{\text{cmd}}, \text{phase})$. 
% Note that only the swing foot is updated, while the stance foot remains fixed to ground until the next transition. 
This keeps nominal stepping physics explicit without task-specific hard constraints. Here $(x,y)$ and $(\dot{x},\dot{y})$ denote the current CoM position and velocity, and $\Pi(\cdot)$ maps the ICP, the commanded velocity $(v_{\text{cmd}}, \dot{\psi}_{\text{cmd}})$, and the gait phase to the foothold target $u_{\text{init}}$.

% \paragraph*{Stage–I on flat ground: disturbance–target training.} Stage–I trains the low-level policy to track template footsteps and remain reliable under small goal variations. During training, we inject a bounded perturbation $\tilde{u}_{\mathrm{init}}=u_{\mathrm{init}}+\varepsilon$, where $\varepsilon\!\sim\!\mathcal{D}\subset\mathbb{R}^3$ with components $(\Delta x,\Delta y,\Delta\psi)$ in the body frame; the perturbation is applied only to the swing-foot landing target and is held fixed until touchdown. The policy observes proprioception (IMU, joint states), step phase features, and the current left/right step commands, and outputs joint position targets tracked by a PD layer at high rate. Rewards emphasize accurate footstep realization and stable contact scheduling, with light regularization for smoothness and safety, as summarized below.

\subsubsection{Foothold tracker training on flat ground} 
Stage-I trains the low-level policy to track footholds and remain reliable under small target disturbances. During training, a bounded perturbation is injected as $\tilde{u}_{\mathrm{init}}=u_{\mathrm{init}}+\varepsilon$, where $\varepsilon=(\delta x,\delta y,\delta\psi)\subset\mathbb{R}^3$ is defined in the body frame as
\begin{equation}
\begin{aligned}
\varepsilon \sim \mathcal{D}
=\mathrm{Unif}\!\big([-\delta_x,\delta_x]\times[-\delta_y,\delta_y]\times[-\delta_\psi,\delta_\psi]\big),\\
\delta_x=\delta_y=0.05~\text{m},\qquad
\delta_\psi=20^\circ~(\approx 0.349~\text{rad})
\end{aligned}
\end{equation}
Note that the perturbation is applied only to the swing foot and is held constant until touchdown. Overall, the policy observes proprioception information (IMU signals and joint states) and current step phase, and outputs joint position commands to the joint PD controller. The reward function emphasizes accurate foothold realization and stable contact scheduling, with light regularization for smoothness and safety. The reward terms are listed in Table~\ref{tab:stage1-rewards} in the appendices, of which the key ones are summarized below.

\emph{(i) step\_tracking}: We reward correct stance leg alternation and precise swing leg foot placement at touchdown as
\begin{equation}
\begin{aligned}
r_{\mathrm{sched}} ={}& w_{\mathrm{alt}}\,(\mathbb{I}_{R}-\mathbb{I}_{L})\,s \\
&+\, w_{\mathrm{pos}}\,\phi_{1}\!\big(\,\|\mathbf p_{\mathrm{sw}}-\mathbf p_{\mathrm{tgt}}\|\,;\,a_p\big) \\
&+\, w_{\mathrm{yaw}}\,\phi_{1}\!\big(\,|\psi_{\mathrm{sw}}-\psi_{\mathrm{tgt}}|\,;\,a_\psi\big)
\end{aligned}
\end{equation}
where $\mathbb{I}_{R}, \mathbb{I}_{L}\!\in\!\{0,1\}$ are right and left foot contact indicators at touchdown, $s\!\in\!\{-1,+1\}$ is the sign for stance leg alternation based on step phase, $\mathbf p_{\mathrm{sw}}$ and $\mathbf p_{\mathrm{tgt}}$ are the actual and desired foothold positions, $\psi_{\mathrm{sw}}$ and $\psi_{\mathrm{tgt}}$ are the actual and desired foothold orientations, and $\phi_{1}$ is defined in Table~\ref{tab:reward-notation-compact} in the appendices. The corresponding scalings used in this term are $w_{\mathrm{alt}}{=}1$, $w_{\mathrm{pos}}{=}5$, $a_p{=}1$, $w_{\mathrm{yaw}}{=}0.5$ and $a_\psi{=}1$. 

\emph{(ii) tracking\_lin\_vel\_world}: We penalize the error between commanded and measured base linear velocities in the world frame (normalized by the command magnitude), fostering faithful velocity following.  

\emph{(iii) base\_heading \& base\_z\_orientation}: We align the base heading angle to the commanded value, and penalize base tilt indicated by projected gravity, stabilizing the base orientation for clean foot placement. 

\emph{(iv) joint\_regularization}: We add a soft regularizer on hip and waist yaw angles and leg abduction/adduction angles to keep them near neutral, avoiding extreme poses during locomotion.

% \paragraph*{Timing and interfaces.}
% The template is updated only at step transition, while the low-level controller runs continuously at the control rate. At each transition, the swing footstep target is computed once and then held constant until touchdown. During each half gait cycle, one leg swings toward this fixed target, whereas the stance foot remains fixed until the next transition. We use the same observation composition and command interface at deployment, preserving sim–real consistency and carrying the Stage–II narrow-path performance from simulation to hardware.

% \subsection{Residual Planner on Beam with Minimal Perception}
\subsection{Stage-II Training for Safe Foothold Modifier}

Stage-II trains a high-level exteroception-based foothold modifier on narrow paths that refines the template-based foothold target from Stage-I. The objective is to prioritize safe and precise foot placement under narrow support, without any manual lateral locking.

% \paragraph*{Minimal, forward-facing elevation cues.}
% To keep sensing lightweight and deployment-friendly, the planner consumes a compact robot-centric elevation window aligned with the body frame (no heavy vision stack). Concretely, we sample a fixed window in front of the robot and flatten it into a vector that complements proprioception: 
% \begin{itemize}
% \item \textbf{Frame and coverage:} body frame with $x$ forward, $y$ lateral; forward range $x\in[0.1,1.1]$\,m, lateral range $y\in[-0.8,0.8]$\,m.
% \item \textbf{Resolution and size:} uniform $0.1$\,m spacing along both axes, yielding an $11\times17$ grid (187 points) when endpoints are included.
% \item \textbf{Flattening convention:} column-major from near-to-far and left-to-right (fixed at training and deployment).
% \end{itemize}
% In simulation, grid points are queried from the terrain height field; on hardware, a LiDAR-based mapper produces the same window in the robot frame. The specification, sampling, and flattening are \emph{identical} across sim and real, ensuring representation consistency.

\subsubsection{Anterior terrain height map}
To keep sensing lightweight and deployment-friendly, the foothold modifier consumes a compact anterior terrain height map aligned with the body frame. The body frame is defined with $x$ pointing forward and $y$ pointing leftward. The terrain height map is sampled within a fixed area in front of the robot and flattened into a vector in the order of from near to far and from left to right. The dimension of this fixed area is designed to be $x\in[0.1,1.1]\,m$ and $y\in[-0.8,0.8]\,m$. The sampling resolution is uniformly $0.1\,m$ along both axes, resulting in $11\times17$ grid points. In simulation, height of these grid points are queried from the terrain height field. In physical experiments, a LiDAR-based mapper produces the same map to ensure consistency.

% \paragraph*{Residual interface on the swing foot.}
\subsubsection{Foothold modifier training on narrow paths}

% At each step transition, the planner outputs a body-frame residual $\Delta r=(\Delta x,\Delta y,\Delta\psi)\in\mathbb{R}^3$ for the \emph{swing} foot only. The final target for foot $i\in\{\text{L},\text{R}\}$ is
% \begin{equation}
% \label{eq:residual-compose}
% \begin{aligned}
% u_{\text{final}}^{(i)} \;=\;
% \begin{cases}
% u_{\text{init}}^{(i)} \oplus \,\text{sat}_{S}\!\big(\Delta r\big), & \text{if } i \text{ is swing},\\
% u_{\text{init}}^{(i)}, & \text{if } i \text{ is stance},
% \end{cases}
% \\[2pt]
% \text{sat}_{S}(\Delta r)=\operatorname{clip}(\Delta r,-S,S),\qquad \Delta r=(\Delta x,\Delta y,\Delta\psi).
% \end{aligned}
% \end{equation}
% where $\oplus$ denotes pose composition in the footstep space and $S{=}(s_x,s_y,s_\psi)$ sets small component-wise bounds. The planner updates around phase transitions (slow timescale), while the low-level tracker runs at the control rate (fast timescale).

At each step transition, the foothold modifier outputs a body-frame residual $\Delta u=(\Delta x,\Delta y,\Delta\psi)\in\mathbb{R}^3$ for the swing foot only. The final foothold target for foot $i\in\{\text{L},\text{R}\}$ is calculated as
\begin{equation}
\label{eq:residual-compose}
\begin{aligned}
&u_{\mathrm{final}}^{(i)} \;=\;
\begin{cases}
u_{\mathrm{init}}^{(i)} \oplus \,\text{sat}_{S}\!\big(\Delta u\big), & \text{if } i \text{ is swing},\\
u_{\mathrm{init}}^{(i)}, & \text{if } i \text{ is stance},
\end{cases}
\\
&\text{sat}_{S}(\Delta u)=\operatorname{clip}(\Delta u,-S,S)%,\qquad \Delta u=(\Delta x,\Delta y,\Delta\psi).
\end{aligned}
\end{equation}
where $\oplus$ denotes pose composition in the foot task space and $S{=}(s_x,s_y,s_\psi)$ sets component-wise foothold adjustment bounds. 
% The planner updates around phase transitions (slow timescale), while the low-level tracker runs at the control rate (fast timescale).

To explicitly reflect locomotion safety and preserve the interpretability from Stage-I policy, the reward terms for Stage-II training are designed as listed in Table~\ref{tab:stage2-rewards}, of which the key ones are summarized below.

\emph{(i) foothold\_safety}: We encourage foothold targets that are within the narrow path and locally flat. We penalize situations include (i) “abyss”, where terrain height is below a safe threshold, and (ii) lack of local flatness around the target foothold, expressed as
\begin{equation}
\begin{aligned}
r_{\mathrm{footstep\_safety}}
&= -\,\mathbb{I}\,\{\,h(\mathbf u_{\mathrm{final}}) < u_{\mathrm{th}}\,\} \\
&\quad -\,\operatorname{Var}\,\big\{\,h(\mathbf u)\;:\;\mathbf u\!\in\!\mathcal N(\mathbf u_{\mathrm{final}})\big\}.
\end{aligned}
\end{equation}
where $h(\mathbf u)$ outputs the terrain height at location $\mathbf u$, $\mathbf u_{\mathrm{final}}$ is the foothold target after modification, $\mathcal N(\mathbf u_{\mathrm{final}})$ represents a small grid patch with $\mathbf u_{\mathrm{final}}$ at its center for assessing local flatness, and $u_{\mathrm{th}}$ is a safety threshold for identifying valid foothold regions.

\begin{figure}[!t] 
    \centering
    \includegraphics[width=\columnwidth]{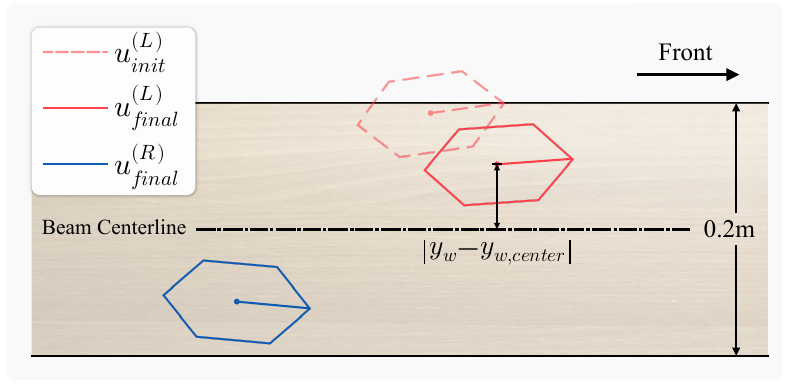}
    % \vspace{-1pt}
    % \caption{The dashed outline shows the template left-foot target $u_{\mathrm{init}}^{(L)}$, while solid polygons denote the finalized contacts $u_{\mathrm{final}}^{(L)}$ (red) and $u_{\mathrm{final}}^{(R)}$ (blue). The vertical marker illustrates the lateral offset of the base from the beam centerline, $|\,y_w - y_{w,\mathrm{center}}\,|$. A smooth Gaussian penalty in this distance encourages staying near the centerline; the beam width is $0.20$\,m.}
    \caption{Demonstration of foothold modification in the Stage-II training. The dashed polygon shows the initial foothold target from the foothold planner for the left leg, $u_{\mathrm{init}}^{(L)}$, while the solid polygons denote the final foothold targets from the foothold modifier for both legs, $u_{\mathrm{final}}^{(L)}$ (red) and $u_{\mathrm{final}}^{(R)}$ (blue). A smooth Gaussian penalty is given to the foothold's lateral offset from the beam centerline, $|\,y_w - y_{w,\mathrm{center}}\,|$, to encourage staying near the centerline}
    \label{fig:footstep}
    % \vspace{-15pt}
\end{figure}

\emph{(ii) beam\_balance}: We encourage centerline adherence by applying a Gaussian shaping to the foothold's lateral deviation, resulting in higher reward for smaller deviation (Fig.~\ref{fig:footstep}).

\emph{(iii) forward\_progress}: We reward forward movement along the narrow path only and discourages the other way.

\emph{(iv) face\_forward}: We encourage alignment between foot orientation and forward direction by applying a shaping function to the foothold's yaw angle, so that a smaller yaw angle yields a larger reward.

\emph{(v) feet\_proximity}: We penalize excessively small distance between the feet along the direction of narrow path to avoid leg interference.

\emph{(vi) action\_magnitude \& action\_smoothness}: We penalize the magnitude of foothold residual and its step-to-step variation to keep the refinement minimal and smooth.

During training, the observation of the high-level modifier policy concatenates: (i) the proprioception information including the IMU signals and joint states, (ii) the step phase features, (iii) current foothold target from the template-based planner, and (iv) the flattened anterior terrain height map. 
% The modifier is queried once per step with its output $\Delta u$ applied to the upcoming swing foot according to~\eqref{eq:residual-compose}. 
Notably, even though no centerline locking or lateral offset is manually added, safe and precise footholds emerge from the compact terrain perception based locomotion control.

\section{Experiments}
\subsection{Setup}
\label{sec:Evaluation Setup}

% We evaluate our approach in simulation and on a Unitree~G1 humanoid. The evaluation protocol, sensing interface, and controller architecture follow the method in Sec.~\ref{sec:method} to ensure apples-to-apples comparison across settings.

The proposed approach is evaluated first in simulation and then on a Unitree G1 humanoid robot. Following Section~\ref{sec:method}, the settings are kept the same between simulation and experiments to ensure apples-to-apples comparison. Both the low-level and high-level policies are exported as TorchScript and executed on the onboard computer in Unitree G1. Standard safety guards, including torque limits, fall/edge detectors and foot–to-foot clearance checks, are enabled during all trials. 

% \paragraph*{Robot and control.}
% Experiments use a Unitree~G1 with standard joint torque limits and PD tracking. The low-level tracker runs at a high control rate (fast loop), while the residual planner is queried around step transitions (slow loop); both policies are exported as TorchScript and executed on an onboard computer. Safety guards are enabled during all trials (torque saturation, fall/edge detectors, foot–foot clearance checks).

% \paragraph*{Sensing and sim–real consistency.}
% The planner consumes exactly the same compact, forward-facing elevation window and proprioception (IMU/joint states) as in Sec.~\ref{sec:method}, with the same coordinate conventions, window coverage, resolution, and flattening order. No additional visual perception is used.

% \paragraph*{Beam scenarios and tasks.}
% In simulation we consider straight beams of widths $\{0.15,\,0.20,\,0.25\}$\,m and lengths $3$–$5$\,m. The primary hardware task is traversal of a $0.20$\,m-wide, $3$\,m-long beam placed on level ground. For each setting we run 20 independent trials from a standardized starting pose and command.

% \paragraph*{Protocol and success criteria.}
% A trial is deemed \emph{successful} if the robot reaches the beam end within a time/step limit, with no falls, and all footfalls remain within beam boundaries (footprint center inside the beam polygon). A trial is a \emph{failure} if any footfall exceeds the beam edge, the torso violates attitude limits, or a protective stop is triggered (e.g., joint-position limit breach or joint-velocity overspeed).

As with narrow paths, straight beams with width of $\{0.15,\,0.20,\,0.25\}\,m$ and length between $3$–$5\,m$ are used during training in simulation. In physical experiments, the narrow path is set to be a wooden beam of $0.2\,m$ wide and $3\,m$ long placed on level ground. For each setting, $20$ independent trials were run with a standardized initial pose and command. For both simulated and physical robot, a trial is deemed successful if the robot reaches the beam end within a time/step limit, with all footholds' centers remain within the beam boundaries and no falls, while a trial is considered a failure if any foothold's center exceeds the beam edge, the torso violates the attitude limits, or a protective stop is triggered by excessive joint position or velocity.

\subsection{Simulation Evaluation}

% We evaluate the proposed \emph{Template + Residual} approach in controlled simulation against strong baselines, and through a focused ablation. Unless otherwise stated, settings follow Sec.~\ref{sec:Evaluation Setup} and the method in Sec.~\ref{sec:method}; training budgets, evaluation episode counts, and seeds are matched across methods.

The proposed approach is evaluated in simulation against two baselines and through one ablation study. 

% \paragraph*{Baselines.}
% \textbf{Template-only}: The robot executes footsteps from the LIP template with no residual refinement at test time. Importantly, the low-level tracker is the same \emph{Stage-I robust tracker trained with disturbance-targets on flat ground} as in our full method; observations and control rates are identical. This baseline isolates the benefit of the high-level residual planner. \textbf{Monolithic RL}: To isolate the effect of the template--residual interface, we include a single-policy baseline that \emph{does not} use any footstep template or residual. The policy receives exactly the same observation as ours (IMU/joints, gait phase, and the forward elevation window) and directly outputs joint position targets at the control rate. Training uses the same beam-aware objectives as Stage-II; any terms that reference a template target are omitted. We match training budgets, evaluation episode counts, decision frequencies, and random seeds to our method for a fair comparison. \textbf{Ours}: the proposed template with a swing-foot residual $(\Delta x,\Delta y,\Delta\psi)$ and the Stage-I robust tracker.

\subsubsection*{Baselines} (i) \textbf{No-Modifier}: This baseline tracks the foothold from the LIPM-based planner with no residual refinement. Consequently, the controller is the same as the Stage-I policy in the proposed full method. This baseline tests the benefit of the high-level modifier. (ii) \textbf{RL-Only}: This baseline further ditches the reduced-order model based foothold planner and utilize an end-to-end RL framework for locomotion control. The learned policy receives exactly the same observation as the proposed method but directly outputs desired joint positions at the $100\,Hz$ control rate, which are then realized by the joint PD controller at $1\,kHz$. The training uses the same reward function as that in the Stage-II training of the proposed method, except the terms that refer to the template-based foothold target are omitted. Besides, training budgets, evaluation episode counts and random seeds are kept the same to yield a fair comparison. 

% \paragraph*{Ablation.}
% We compare \emph{Ours (full)}—the low-level tracker trained on flat ground with small random perturbations added to LIP footsteps—against \emph{w/o disturbances}, which is trained identically but \emph{without} these perturbations. All other factors (observations, rewards, network size, optimization hyperparameters, training budget, seeds, and the subsequent Stage–II training) are held constant. This ablation tests whether disturbance–target training yields a tracker that better tolerates target variations at step transitions, improving success rate, centerline deviation, and foot-placement RMSE on beam tasks.

\subsubsection*{Ablation study}

The effect of the small random perturbations added to the LIPM-based foothold targets in the Stage-I training are studied through ablation. The resulted policy is trained identically to the proposed method except without these perturbations. 
% All the other settings, including observations, rewards, network size, optimization hyperparameters, training budget, random seeds and the subsequent Stage-II training, are kept the same. 
This ablation study tests whether target disturbance during training yields a foothold tracker that better tolerates target variations at step transitions, thus improving foot-placement RMSE, centerline following and success rate on narrow path traversal tasks.

% \paragraph*{Metrics and statistics.}
% We report three metrics: \textbf{success rate} (\%), \textbf{centerline deviation} (m), and \textbf{foot-placement RMSE} (m). Each configuration is evaluated over 20 episodes. Aggregate results for the main comparison and the focused ablations are summarized in Tables~\ref{tab:sim-main} and \ref{tab:sim-ablate}.

The evaluation in simulation uses three metrics: success rate (\%), centerline deviation (m), and foot-placement RMSE (m). Each metric is evaluated over 20 episodes. The results for the baseline comparisons and the ablation study are summarized in Tables~\ref{tab:sim-main} and \ref{tab:sim-ablate}, respectively. It can be seen that, across different path widths and lengths, the proposed method can improve success rate and reduce centerline deviation compared to the other two baselines. As with the foot-placement RMSE, it is computed with respect to the commanded foothold target $u_{\mathrm{cmd}}$. For the \emph{No-Modifier} baseline, perturbation from the modifier's residual is disabled at test time, so that $u_{\mathrm{cmd}}{=}u_{\mathrm{init}}$ and the tracker can follow clean foothold targets with small RMSE. However, the commanded foothold target in our proposed method is calculated as $u_{\mathrm{cmd}}{=}u_{\mathrm{final}}=u_{\mathrm{init}}\oplus \Delta u$, with perturbation from the nonzero residual $\Delta u$. Therefore, realizing these targets leads to a slightly larger foot-placement RMSE.
On the other hand, results from the ablation study indicate that removing Stage-I foothold disturbances leads to poorer tracking performance at step transition and larger foot-placement errors.

\begin{table}[!htbp]
\centering
\footnotesize
\begin{threeparttable}
\caption{Baseline comparison results from walking on beams in simulation (mean~$\pm$~std over 20 runs)}
\label{tab:sim-main}
\setlength{\tabcolsep}{1pt}
\begin{tabular*}{\columnwidth}{@{\extracolsep{\fill}}lccc@{}}
\toprule
\textbf{Method} & \shortstack{\textbf{Success}\\\textbf{rate (\%)}} & \shortstack{\textbf{Centerline}\\\textbf{dev. (m)}} & \shortstack{\textbf{FP-RMSE}\\\textbf{(m)}} \\
\midrule
No-Modifier & 15 & 0.04690~$\pm$~0.00057 & 0.01962~$\pm$~0.00079 \\
RL-Only & 10 & 0.18192~$\pm$~0.07075 & \textemdash\tnote{a} \\
\textbf{Ours} & \textbf{100} & \textbf{0.01639~$\pm$~0.00117} & \textbf{0.02633~$\pm$~0.00083} \\
\bottomrule
\end{tabular*}
\begin{tablenotes}[flushleft]
\footnotesize
\item[a] Not applicable: RL-Only outputs joint positions directly and does not provide commanded foothold targets for this metric.
\end{tablenotes}
\end{threeparttable}
\end{table}

\begin{table}[!htbp]
\centering
\footnotesize
\setlength{\tabcolsep}{1.5pt}
\caption{Ablation study results from walking on a $0.20\,m$-wide beam in simulation (mean~$\pm$~std over 20 runs)}
\label{tab:sim-ablate}
\begin{tabular*}{\columnwidth}{@{\extracolsep{\fill}}lccc@{}}
\toprule
\textbf{Configuration} & \shortstack{\textbf{Success}\\\textbf{rate (\%)}} & \shortstack{\textbf{Centerline}\\\textbf{dev. (m)}} & \shortstack{\textbf{FP-RMSE}\\\textbf{(m)}} \\
\midrule
\shortstack[l]{w/o Stage-I\\disturbances} & 50 & 0.09696~$\pm$~0.07876 & 0.05467~$\pm$~0.05467 \\
\textbf{Ours} & \textbf{100} & \textbf{0.01639~$\pm$~0.00117} & \textbf{0.02633~$\pm$~0.00083} \\
\bottomrule
\end{tabular*}
\end{table}

% \paragraph*{Results overview.}
% Across beam widths, \textbf{Ours} improves success rate and reduces centerline deviation compared to \textbf{Template-only} and \textbf{Monolithic RL}. Removing Stage-I disturbances yields brittle tracking at step switches and larger placement errors.

% \paragraph*{Error attribution (simulation).}
% One dominant failure modes is observed: pronounced lateral/heading oscillations shrink the feasible on-path set in the local elevation window. The template target can drift toward the edge, and the planner drives the residual toward its bounds ($\|r\|\!\to\!S$) yet still cannot pull the target back onto the path, leading to growing centerline deviation and an eventual off-path footfall.

It is worth noting that one dominant failure mode can be observed in the process of training. When pronounced heading angle oscillation occurs on the robot, the feasible foothold set in the local terrain height map shrinks. Consequently, the initial foothold target from the planner policy can drift toward the path edge and drive the residual from the modifier policy to reach its bounds ($\|r\|\!\to\!S$), yet still unable to pull the final foothold target back onto the narrow path. This leads to a failure with off-path footholds.

\subsection{Hardware Validation}

% We validate the full system on a Unitree~G1 traversing a $0.20$\,m-wide, $3$\,m-long beam placed on level ground. The controller stack, action/observation interfaces, and sensing representation strictly match Sec.~\ref{sec:method}. Each configuration is tested for $N$ independent trials from a standardized start.

% \paragraph*{LiDAR elevation window on hardware.}
% A body-centric elevation window is constructed online from a LiDAR stream using the same coverage and resolution as training (Sec.~\ref{sec:method}): $x\!\in\![0.1,1.1]$\,m forward, $y\!\in\![-0.8,0.8]$\,m lateral, $\Delta x\!=\!\Delta y\!=\!0.1$\,m ($11{\times}17$ grid). The pipeline is: (i) transform raw points using IMU gravity alignment into the robot/body frame and crop to the fixed region of interest (ROI); (ii) grid the ROI at 0.1\,m resolution and, for each cell, use adaptive square binning (start 0.1\,m, expand by 0.05\,m up to 0.3\,m until nonempty) and take the \emph{maximum} height \(z_{\max}\) of the in-cell points as the cell estimate; (iii) add a fixed \(z\)-offset of \(+0.38\)\,m and clamp heights to \([-1.4,\,-0.7]\)\,m to mitigate sparse returns and sensor bias; (iv) reshape to an \(11\times17\) grid and flatten in \emph{column-major} order (near\(\rightarrow\)far within each lateral column; columns left\(\rightarrow\)right), publishing at a fixed rate synchronized with planner queries. This representation is identical to the simulation input in coverage, resolution, coordinate conventions, and flattening order.

As mentioned in the setup, the full proposed method is validated on a Unitree~G1 humanoid robot. $20$ independent trials were carried out. During each trial, a body-centric anterior terrain height map with the same dimension and resolution as in training is constructed online using the onboard LiDAR stream, as shown in Fig.~\ref{fig:lidar}. Representative simulation and hardware traversal results are shown in Supplementary Movie~1. The pipeline for building the height map is: (i) transform raw data points into the robot's body frame using the IMU's gravity perception and crop the data to the fixed region of interest (ROI), (ii) grid the ROI at $0.1\,m$ resolution, use an adaptive binning square for each grid point (the adaptive binning square starts with $0.1\,m$ side length and expand up to $0.3\,m$ by $0.05\,m$ until nonempty), and take the maximum height $z_{\max}$ inside the square as the estimated height, (iii) clamp the heights to $[0.7,\,1.4]\,m$ to mitigate sparse returns and sensor bias, and (iv) flatten the $11\times17$ height map to a vector and publish at a fixed rate synchronized with the modifier queries. 

\begin{figure}[!t] 
    \centering
    \includegraphics[width=\columnwidth]{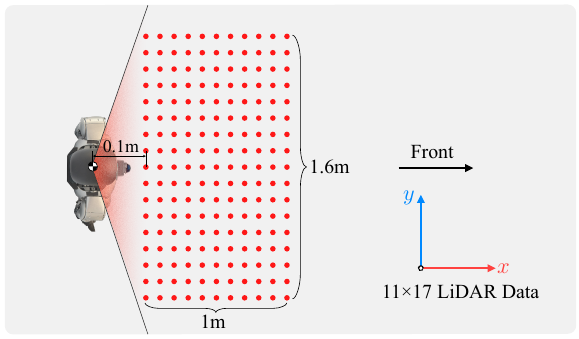}
    % \vspace{-1pt}
    % \caption{Top-down view of the Minimal Perception Window, which captures a forward-facing elevation grid from the robot's LiDAR sensor. The window spans $x \in [0.1, 1.1]$ m and $y \in [-0.8, 0.8]$ m, with a resolution of 0.1 m, providing a compact, real-time terrain representation for footstep planning.}
    \caption{Top view of the anterior terrain height map based on the robot's LiDAR sensor. The map spans $x \in [0.1, 1.1]\,m$ and $y \in [-0.8, 0.8]\,m$ with a resolution of $0.1\,m$ uniformly, providing a real-time terrain representation for foothold modification}
    \label{fig:lidar}
    % \vspace{-15pt}
\end{figure}

% \paragraph*{Protocol and safety.}
% Both policies are executed as TorchScript on the onboard computer; the residual planner updates around step transitions, and the low-level tracker runs at the control rate. Safety guards (torque saturation, fall/edge detectors, and foot–foot clearance checks) remain active for all trials.

% \paragraph*{Results.}
% We report two task-level metrics on the $0.20$\,m\,$\times$\,3\,m beam: (i) \emph{success rate} (percentage of trials that reach the end under constraints), and (ii) \emph{traversal rate}, defined as the fraction of the beam length completed before failure (per trial $r_i=\min(1, d_i/L_{\text{beam}})$, averaged over trials). For a head-to-head comparison, we include the SOTA \textbf{BeamDojo}\cite{BeamDojo2025} baseline evaluated on the same real-beam setting. Our method attains high success with consistent progress along the beam; the detailed results are summarized in Table~\ref{tab:hw-main}.

Two task-level metrics are used to evaluate the robot's experimental performance on the $3\,m\times0.2\,m$ beam. Besides the success rate used in simulation evaluation, traversal rate is selected and defined as the fraction of the beam length completed before failure. The traversal rate is calculated as $r_i=\min(1, d_i/L_{\text{beam}})$ per trial and then averaged over all trials. For a head-to-head comparison, the state-of-the-art results from \textbf{BeamDojo}~\cite{BeamDojo2025} are referred to as the baseline for the experiments. It can be seen from Table~\ref{tab:hw-main} that the proposed method in this paper obtains higher success rate and traversal rate along the beam.

\begin{table}[!htbp]
\centering
\footnotesize
\caption{Hardware comparison results from walking on a $3\,m\,\times\,0.2\,m$ beam in reality (mean~$\pm$~std). 
Ours: $N{=}20$ trials; BeamDojo: $N{=}5$ trials as reported}
\label{tab:hw-main}
\begin{tabular*}{\columnwidth}{@{\extracolsep{\fill}}lcc@{}}
\toprule
\textbf{Setting} & \shortstack{\textbf{Success}\\\textbf{rate (\%)}} & \shortstack{\textbf{Traversal}\\\textbf{rate (\%)}} \\
\midrule
BeamDojo \cite{BeamDojo2025} (G1, real beam) & 80 & 88.16 \\
\textbf{Ours (G1, real beam)} & \textbf{100} & \textbf{100} \\
\bottomrule
\end{tabular*}
\end{table}

% \paragraph*{Error attribution (hardware).}
% During bring-up and pilot runs, we occasionally observed \emph{height-estimation bias near the path boundary}: sparse/reflective LiDAR returns at the edge caused under/over-estimates in a few cells and could nudge the selected foothold toward the boundary. \textbf{Mitigation:} robust per-cell statistics (median with outlier rejection) and mild temporal smoothing of the elevation window, tighter ROI alignment, conservative residual bounds, and slightly increased safety weights at deployment. After these changes, the evaluation runs reported in Table~\ref{tab:hw-main} did not exhibit this failure.

During bring-up and pilot runs in experiments, height-estimation bias near the path boundary was occasionally observed, which could nudge the foothold target toward the boundary. To mitigate this defect, several measures were performed, including applying robust per-grid statistics (median with outlier rejection), mild temporal smoothing of the height map and conservative residual bounds. Consequently, the evaluation runs reported in Table~\ref{tab:hw-main} did not exhibit this failure any more.

\section{Conclusion}
% \textbf{Key insights.}
% Our results indicate that (i) adding a small residual over a LIP template improves success rate, centerline adherence, and safety margins while keeping the residual small and smooth; (ii) Stage–I disturbance–target training is crucial—removing it increases touchdown errors and off–beam contacts; (iii) body–frame $(\Delta x,\Delta y,\Delta\psi)$ provides a more stationary, task–aligned interface than world–frame residuals.

This paper proposes a two-stage reinforcement learning framework for robust and safe humanoid walking control when traversing narrow paths. The key insights indicated by the results include  (i) foothold target disturbances added in Stage-I is crucial to reducing touchdown errors and off–beam steps and (ii) a small and smooth foothold residual added in Stage-II to the LIPM-based foothold improves success rate, centerline adherence and safety margins when traversing narrow paths. Besides, a compact anterior terrain height map is sufficient for foothold decisions and sim–to–real transfer simplification. No heavy vision pipelines are needed.

In the future, we will (i) extend the proposed framework beyond straight beams to more sparse-support terrains (e.g., stepping stones, gaps, curved beams), and (ii) expand the foothold representation to 3D with a nominal vertical profile $z^{*}$ to enable traversing stairs and uneven terrains.

\appendix
\subsection{Reward Functions}
The reward functions used during the two-stage training process are shown
in Tables~\ref{tab:stage1-rewards} and \ref{tab:stage2-rewards}. The corresponding symbols and their definitions are provided
in Table~\ref{tab:reward-notation-compact}.

\begin{table}[!htbp]
\centering
\footnotesize
\setlength{\tabcolsep}{2pt}
\caption{Stage-I rewards}
\label{tab:stage1-rewards}
\begin{tabular*}{\columnwidth}{@{\extracolsep{\fill}}lcl@{}}
\toprule
\textbf{Term} & \textbf{Weight} & \textbf{Equation} \\
\midrule
step\_tracking & $3.0$ &
$\begin{aligned}[t]
&\big(\mathbb{I}_R-\mathbb{I}_L\big)\,s \;\times\; \phi_1(\|\delta_p\|_2;\ a_p\!=\!1)\\
&\times\; \phi_1(|\delta_\psi|;\ a_r\!=\!1)
\end{aligned}$ \\
\shortstack[l]{tracking\_lin\_\\vel\_world} & $4.0$ & 
$\begin{aligned}[t]
&\phi_2\!\Big(\!\big\|\,(v^c_{xy}-v_{xy})\odot{}\\
&\qquad (1+|v^c_{xy}|)^{-1}\big\|_2;\ a_v\!=\!1\Big)
\end{aligned}$ \\
base\_heading & $3.0$ &
$\displaystyle \phi_1\!\big(\,|\mathrm{wrap}(\psi^c-\psi)|;\ a_\psi\!=\!\tfrac{\pi}{2}\big)$ \\
\shortstack[l]{base\_z\_\\orientation} & $1.0$ &
$\displaystyle \phi_2\!\big(\,\|g_{xy}\|_2;\ a_g\!=\!0.2\big)$ \\
base\_height & $1.0$ &
$\displaystyle \phi_2\!\big(\,h-h^\ast;\ a_h\!=\!1\big)$ \\
\shortstack[l]{joint\_\\regularization} & $1.0$ &
$\displaystyle \tfrac{1}{|\mathcal J|}\!\sum_{j\in\mathcal J}\phi_2(q_j;\ a_q\!=\!1)$ \\
lin\_vel\_z & $1{\times}10^{-1}$ & $-\,v_z^2$ \\
ang\_vel\_xy & $1{\times}10^{-2}$ & $-\,\|\omega_{xy}\|_2^2$ \\
dof\_vel & $1{\times}10^{-3}$ & $-\,\|\dot{\boldsymbol q}\|_2^2$ \\
torques & $1{\times}10^{-4}$ & $-\,\|\boldsymbol\tau\|_2^2$ \\
actuation\_rate & $1{\times}10^{-3}$ &
$\displaystyle -\,\|\boldsymbol a_t-\boldsymbol a_{t-1}\|_2^2/\Delta t^{\,2}$ \\
actuation\_rate2 & $1{\times}10^{-4}$ &
$\displaystyle -\,\|\boldsymbol a_t-2\boldsymbol a_{t-1}+\boldsymbol a_{t-2}\|_2^2/\Delta t^{\,2}$ \\
dof\_pos\_limits & $10$ &
$\displaystyle -\!\sum_j\!\big[(\ell_j-q_j)_+ + (q_j-u_j)_+\big]$ \\
torque\_limits & $1{\times}10^{-2}$ &
$\displaystyle -\!\sum_j\!\big(|\tau_j|-0.8\,\tau^{\max}_j\big)_+$ \\
\bottomrule
\end{tabular*}
\end{table}

\begin{table}[!htbp]
\centering
\footnotesize
\setlength{\tabcolsep}{2pt}
\caption{Stage-II rewards}
\label{tab:stage2-rewards}
\begin{tabular*}{\columnwidth}{@{\extracolsep{\fill}}lcl@{}}
\toprule
\textbf{Term} & \textbf{Weight} & \textbf{Equation} \\
\midrule
foothold\_safety & $1.0$ &
$\displaystyle -\,5\!\!\sum_{f\in\{\mathrm L,\mathrm R\}}\! \mathbb{I}\!\{\,h(\mathbf p_t^{\,f})<\!-0.20\,\}\, m^{\,f}_{\mathrm{swing}}$ \\
beam\_balance & $1.0$ &
$\displaystyle \exp\!\big(\!-\,(|y-y_c|/\sigma_y)^2\big) - 1$ \\
feet\_proximity & $0.1$ &
$\displaystyle -\,\frac{(d_{\min}-|x_R-x_L|)_+}{d_{\min}}$ \\
forward\_progress & $1.0$ &
$\displaystyle \max\!\big(0,\ x_t-x_{t-1}\big)$ \\
face\_forward & $0.1$ &
$\displaystyle \max\!\big(0,\,1-|\mathrm{wrap}(\psi)|/\pi\big)$ \\
contact\_schedule & $1.0$ &
$\begin{aligned}[t]
&\big(\mathbb{I}_R-\mathbb{I}_L\big)\,s \;\times\; \phi_1(\|\delta_p\|_2;\ a_p\!=\!1)\\
&\times\; \phi_1(|\delta_\psi|;\ a_r\!=\!1)
\end{aligned}$ \\
\shortstack[l]{tracking\_lin\_\\vel\_world} & $2.0$ &
$\begin{aligned}[t]
&\phi_2\!\Big(\!\big\|\,(v^c_{xy}-v_{xy})\odot{}\\
&\qquad (1+|v^c_{xy}|)^{-1}\big\|_2;\ a_v\!=\!1\Big)
\end{aligned}$ \\
base\_heading & $0.2$ &
$\displaystyle \phi_1\!\big(|\mathrm{wrap}(\psi^c-\psi)|;\ a_\psi\!=\!\tfrac{\pi}{2}\big)$ \\
\shortstack[l]{base\_z\_\\orientation} & $0.5$ &
$\displaystyle \phi_2\!\big(\|g_{xy}\|_2;\ a_g\!=\!0.2\big)$ \\
base\_height & $0.2$ &
$\displaystyle \phi_2\!\big(h-h^\ast;\ a_h\!=\!1\big)$ \\
action\_magnitude & $0.01$ & $-\,\|\mathbf r_t\|_2^2$ \\
\shortstack[l]{action\_\\smoothness} & $0.01$ & $-\,\|\mathbf r_t-\mathbf r_{t-1}\|_2^2$ \\
\bottomrule
\end{tabular*}
\end{table}

\begin{table}[!htbp]
\centering
\footnotesize
\setlength{\tabcolsep}{3pt}
\caption{Used symbols and constants for reward tables}
\label{tab:reward-notation-compact}
\begin{tabular}{@{}l l@{}}
\toprule
\textbf{Symbol} & \textbf{Definition / Value} \\
\midrule
$\phi_1(e;a)$ & $\exp\!\big(-|e|/(a\,\sigma)\big)$ \\
$\phi_2(e;a)$ & $\exp\!\big(-(e/a)^2/\sigma\big)$ \\
$\sigma$ & tracking shape scale $=0.25$  \\
$h^\ast$ & base height target $=0.78$\,m \\
$\mathcal J$ & joints $\{0,1,5,6\}$ (hip/waist yaw/ab-ad soft centering) \\
$\delta_p,\ \delta_\psi$ & swing-foot pos./yaw error at touchdown (to target) \\
$s$ & gait schedule sign; $\mathbb{I}_R,\mathbb{I}_L$: contact indicators \\
$y_c$ & beam centerline; $\sigma_y=0.1$\,m \\
$d_{\min}$ & inter-foot distance threshold along $x$, $=0.1$\,m \\
$\mathbf r_t$ & footstep residual $(\Delta x,\Delta y,\Delta\psi)$ \\
$(\cdot)_+$ & $\max(\cdot,0)$; $\Delta t$: control step \\
\bottomrule
\end{tabular}
\end{table}

\subsection{Domain Randomization}

The parameter settings for domain randomization are provided in Table~\ref{tab:dr}.

\begin{table}[!htbp]
\caption{Domain Randomization Setting}
\label{tab:dr}
\centering
\footnotesize
\begin{tabular}{@{}l l@{}}
\toprule
\textbf{Term} & \textbf{Value} \\
\midrule
\multicolumn{2}{l}{\textit{Observations}} \\
\midrule
angular velocity noise & $\mathcal U(-0.2,\;0.2)\ \text{rad/s}$ \\
projected gravity noise & $\mathcal U(-0.05,\;0.05)$ \\
joint position noise & $\mathcal U(-0.01,\;0.01)\ \text{rad}$ \\
joint velocity noise & $\mathcal U(-0.01,\;0.01)\ \text{rad/s}$ \\
height measurement noise & $\mathcal U(-0.10,\;0.10)\ \text{m}$ \\
\midrule
\multicolumn{2}{l}{\textit{Humanoid Physical Properties}} \\
\midrule
payload mass (added mass) & $\mathcal U(-1.0,\;1.0)\ \text{kg}$ \\
external push (interval / max vel) & every $2.5$\,s,\ $\|\mathbf v_{xy}\|\!\le\!0.5$\,m/s \\
\midrule
\multicolumn{2}{l}{\textit{Terrain Dynamics}} \\
\midrule
friction coefficient & $\mathcal U(0.5,\;1.25)$ \\
restitution & fixed $=0$ \\
\midrule
\multicolumn{2}{l}{\textit{Elevation Map}} \\
\midrule
window/grid (sim \& real) & fixed ROI, fixed grid; no DR \\
measurement noise (map) & $\mathcal U(-0.10,\;0.10)\ \text{m}$ \\
\bottomrule
\end{tabular}
\end{table}

\subsection{Hyperparameters}

The hyperparameter values used in the two-stage training process can be found in Tables~\ref{tab:stage1-hparams} and~\ref{tab:stage2-hparams}.

\begin{table}[!htbp]
\caption{Stage-I Training Hyperparameters}
\label{tab:stage1-hparams}
\centering
\footnotesize
\begin{tabular}{@{}l l@{}}
\toprule
\textbf{Term} & \textbf{Value} \\
\midrule
\multicolumn{2}{l}{\textit{Rollout / Runner}} \\
\midrule
parallel envs & $4096$ \\
steps per env & $24$ \\
rollout size / update & $4096 \times 24$ samples \\
max iterations & $5000$ \\
save interval & $100$ \\
episode length & $5$\,s \\
policy / algo class & \texttt{ActorCritic} / \texttt{PPO} \\
\midrule
\multicolumn{2}{l}{\textit{PPO / Optimization}} \\
\midrule
learning rate & $1{\times}10^{-5}$ (\texttt{schedule=adaptive}) \\
num learning epochs & $5$ \\
num mini-batches & $4$ \\
clip range & $0.2$ \\
entropy coef & $0.01$ \\
value loss coef & $1.0$ (clipped value: \texttt{True}) \\
discount $\gamma$ & $0.99$ \\
GAE $\lambda$ & $0.95$ \\
desired KL & $0.01$ \\
max grad norm & $1.0$ \\
\bottomrule
\end{tabular}
\end{table}

\begin{table}[!htbp]
\caption{Stage-II Training Hyperparameters}
\label{tab:stage2-hparams}
\centering
\footnotesize
\begin{tabular}{@{}l l@{}}
\toprule
\textbf{Term} & \textbf{Value} \\
\midrule
\multicolumn{2}{l}{\textit{Rollout / Runner}} \\
\midrule
parallel envs & 1024 \\
steps per env & \emph{inherited (not overridden)} \\
max iterations & $10000$ \\
save interval & $100$ \\
policy init noise & $1.0$  \\
action space & residual $(\Delta x,\Delta y,\Delta\psi)$, dim $=3$ \\
control decimation & $10$ \\
\midrule
\multicolumn{2}{l}{\textit{PPO / Optimization}} \\
\midrule
learning rate & $1{\times}10^{-5}$ (\texttt{schedule=adaptive}) \\
num learning epochs & $5$ \\
num mini-batches & $4$ \\
clip range & $0.2$ \\
entropy coef & $0.01$ \\
discount $\gamma$ & $0.99$ \\
GAE $\lambda$ & $0.95$ \\
desired KL & $0.01$ \\
max grad norm & $1.0$ \\
\bottomrule
\end{tabular}
\end{table}

% Finish appendix floats before the acknowledgment and references.
\FloatBarrier

\section*{Acknowledgment}
The authors gratefully acknowledge Runchen Xu for his assistance in preparing the figures for this manuscript.

% Optional statements retained from the authors' source, not required sections
% in the supplied ICARA template. Confirm applicability before enabling.
% This is not an anonymization step. Disclose actual relevant conflicts wherever
% the conference/submission form requires; do not assume "no conflicts".
% \section*{Author Contributions}
% Tianchen Huang developed the overall framework, implemented the code, conducted policy training, performed the experiments, and wrote the initial draft. Wei Zhou assisted with the experimental validation. Ya Wang, Yumin Wang, and Deguang Lyu conducted data collection and analysis. Yu Wang and Shiwu Zhang contributed to the main technical ideas and research innovation. Wei Gao contributed to the conceptual development of the study and assisted with manuscript writing. All authors reviewed and approved the final manuscript.

% \section*{Conflicts of Interest}
% The authors declare that they have no known competing financial interests or personal relationships that could have appeared to influence the work reported in this paper.

% Manual final-page column break for this exact compiled revision.
% Recheck this trigger after changes to text, floats, acknowledgments, or references.
\IEEEtriggeratref{3}
\bibliographystyle{IEEEtran}
\bibliography{references}

\end{document}